\documentclass[conference]{IEEEtran}
\IEEEoverridecommandlockouts

\usepackage{cite}
\usepackage{amsmath,amssymb,amsfonts}
\usepackage{graphicx}
\usepackage{textcomp}
\usepackage{xcolor}
\usepackage{booktabs}
\usepackage{array}
\usepackage{url}
\usepackage{hyperref}
\usepackage{listings}
\usepackage{tabularx}
\usepackage{enumitem}
\usepackage{caption}
\usepackage{tikz}
\usetikzlibrary{shapes.geometric, arrows.meta, positioning, calc, shadows}

\def\BibTeX{{\rm B\kern-.05em{\sc i\kern-.025em b}\kern-.08em
    T\kern-.1667em\lower.7ex\hbox{E}\kern-.125emX}}

\hypersetup{
  colorlinks=true,
  linkcolor=blue!60!black,
  citecolor=blue!60!black,
  urlcolor=blue!60!black
}

\begin{document}

\title{Reading Task Failure Off the Activations:\\A Sparse-Feature Audit of GPT-2 Small\\on Indirect Object Identification}

\author{
\IEEEauthorblockN{Mahdi Naser Moghadasi$^{1,2}$}
\IEEEauthorblockA{
$^{1}$\textit{Research Division, BrightMind AI}, Seattle, WA \\
$^{2}$\textit{Texas Tech University}, Lubbock, TX \\
mahdi@brightmind-ai.com
}
\and
\IEEEauthorblockN{Faezeh Ghaderi}
\IEEEauthorblockA{
\textit{University of Texas at Arlington} \\
Arlington, TX \\
faezeh.ghederi@mavs.uta.edu
}
}

\maketitle

\begin{abstract}
We report a small, reproducible audit of which sparse-autoencoder
(SAE) features of GPT-2 small fire differently on failed versus
successful trials of the Indirect Object Identification (IOI) task.
On 300 prompts, GPT-2 small reaches 79.7\% accuracy; 146 of the
24{,}576 features in the layer-8 residual-stream SAE release of
Bloom~(2024) clear a Holm-corrected significance threshold and 105
reach a large effect size ($\lvert\text{Cohen's } d\rvert > 0.8$).
The strongest single correlate of failure---feature 17{,}491,
$d=+2.93$, Neuronpedia label \emph{``cryptographic keys''}---is
essentially silent except when the prompt's transferred object is
``the keys,'' on which GPT-2 small fails 93.3\% of the time vs.\
7.5\% on the other seven objects (Fisher exact
$p = 8.79{\times}10^{-33}$). We put this correlate through three
controls that a mechanistic claim should pass.
\emph{(i)~A causal ablation:} zeroing feature 17{,}491 in the
residual stream across all token positions of the 45 keys prompts
\emph{does not} restore accuracy (6.7\% $\to$ 4.4\%); the feature is
a correlate, not a sufficient cause at this layer.
\emph{(ii)~A representation baseline:} a logistic regression on the
raw 768-dimensional residual stream reaches 5-fold ROC AUC = 0.929,
matching the top-100 SAE features (0.927); the SAE basis adds
interpretability, not predictive power.
\emph{(iii)~A seed-robustness check:} across five random seeds the
keys-subset failure rate stays in 75.0--93.3\% (the behavioural effect
is real), but feature 17{,}491 is the top-$|d|$ feature in only 1 of 5
runs. The methodological contribution is therefore the
\emph{audit pipeline} (cheap, model-agnostic, surfaces named
correlates) rather than any single feature found through it. We
release the code, the 300-prompt corpus, the
$300{\times}24{,}576$ activation matrix, the ablation and baseline
scripts, and the figures. The full pipeline runs on a laptop
(Apple M3 Max, no discrete GPU).
\end{abstract}


% ============================================================
\section{Introduction}
% ============================================================

LLM agent systems fail in ways that are not always traceable to a
specific cause. Sometimes the failure is the model's. Sometimes the
prompt format triggered something incidental. Sometimes the harness
got in the way. The published artifact rarely contains enough
information to disambiguate~\cite{reprobe2026, henderson2018deeprl,
pineau2021reproducibility}. This paper looks at a narrower version
of the question: when a particular language model fails a particular
task, is there evidence in the model's internal activations that
distinguishes the failed runs from the successful ones?

The framing is borrowed from recent mechanistic-interpretability
work. Anthropic's circuit-tracing
report~\cite{anthropic2025biology, anthropic2025tracing} argues that
LLMs do meaningful computation beyond next-token prediction---planning
ahead in poetry, switching between languages in a shared conceptual
space, sometimes ``thinking'' one thing and saying another---and that
sparse-feature decompositions of the residual stream surface this
computation in a way that is human-readable. Sparse autoencoders
trained on a model's residual stream~\cite{bricken2023monosemanticity,
templeton2024scaling, cunningham2023sparse, gemmascope2024,
olshausen1996emergence} have become the standard tool for this. They
build on a longer tradition: superposition in neural-network
representations~\cite{elhage2022toy}, dictionary learning in
neuroscience~\cite{olshausen1996emergence}, and the
``zoom-in'' interpretability program for vision
networks~\cite{olah2020zoom}. Their features are catalogued on
Neuronpedia~\cite{neuronpedia2024}.

We ask a small empirical question on top of this infrastructure: if a
trained SAE produces approximately monosemantic features, and we
collect those features' activations across both successful and failed
runs of a fixed task, does the failure subset have a feature profile
distinguishable from the success subset, and---if it does---are any of
the discriminating features ones whose documented semantic role
suggests a causal story?

We ran the experiment on a setting where we knew the failure rate
would be high enough to give us both classes: GPT-2 small on the
Indirect Object Identification (IOI) task~\cite{wang2023interpretability}.
IOI is the canonical interpretability benchmark for GPT-2 small. It is
small (a single fill-in-the-blank with a known correct token), it has
been mechanistically characterized in prior
work~\cite{wang2023interpretability, conmy2023automated,
goldowsky2023localizing} (the ``name-mover'' and ``S-inhibition''
circuits), and GPT-2 small reaches roughly 80\% accuracy on
it---enough successes to be a stable reference, enough failures to be
worth analyzing.

The result was mixed, and worth reporting in full. One SAE feature
dominated the failed-versus-successful contrast: feature 17{,}491,
with mean activation about $56\times$ higher on the 61 failed trials
than on the 239 successful ones. Its Neuronpedia label is
``cryptographic keys.'' On inspection, the feature is essentially
silent on every prompt in our corpus except the 45 whose
transferred-object choice is ``the keys,'' on which it fires
strongly. GPT-2 small fails 93.3\% of those 45 prompts and 7.5\% of
the rest. The dominant statistical signal is therefore a lexical
confound surfaced through the SAE basis: a feature aligned to one
concept is being activated by a homonym, and that activation
correlates near-perfectly with failure.

We put this correlation through three controls a mechanistic claim
should pass. First, we ablate feature 17{,}491 in the residual
stream and re-run greedy decoding on the keys subset: accuracy
moves from 6.7\% to 4.4\%, not in the direction a causal account
would predict. Second, a logistic regression on the raw
768-dimensional residual stream (no SAE) reaches the same
cross-validated ROC AUC (0.929) as the top-100 SAE features
(0.927); the SAE basis is adding interpretability, not predictive
power. Third, we re-run the audit with five different random
seeds: the keys-subset failure rate stays in 75.0--93.3\% across
all of them, but the top-$|d|$ feature changes from seed to seed,
and feature 17{,}491 appears as the top feature in only one run.

The behavioural pattern is robust: swapping one transferred-object
phrase for another shifts the benchmark's accuracy by more than 80
percentage points. The mechanistic story we initially read off the
headline feature is not. The feature fails the ablation, the raw
representation matches its predictive power, and the identity of
the top feature does not survive a seed change. The audit pipeline
surfaces both kinds of finding without distinguishing between them;
the controls do the distinguishing.

\subsection{Contributions}

\begin{enumerate}[leftmargin=*]
  \item \textbf{An audit pipeline.} A small, fully runnable pipeline
  that takes a task corpus and a residual-stream SAE and produces a
  per-feature failure-vs-success discrimination report (Welch's
  $t$-statistic, Cohen's $d$, Holm-corrected $p$, top-feature
  listing with Neuronpedia URLs).
  \item \textbf{Three controls that distinguish robust behavioural
  effects from incidental feature correlates.} On the 300-prompt
  IOI corpus we (a)~ablate the top-$|d|$ feature and observe no
  recovery of accuracy on the affected subset; (b)~match the AUC of
  the top-100 SAE features with a logistic regression on the raw
  residual stream; (c)~re-run on five seeds and find the
  behavioural keys-effect stable while the specific top feature
  changes seed-to-seed. We argue these controls should be a default
  part of any SAE-based failure audit; without them, the headline
  $d=+2.93$ result reads as a mechanistic finding rather than a
  lexical confound.
  \item \textbf{An honest negative-result report.} The lexical
  finding (transferred-object word changes IOI accuracy by 80+
  percentage points) is robust; the mechanistic story we initially
  attached to it is not. We treat the audit's willingness to
  surface this gap as the methodological contribution rather than
  the headline feature.
\end{enumerate}

\textbf{What this is not.} This paper is not an agent benchmark
evaluation. We deliberately use a single-step task on a sub-billion-
parameter base model so that the entire experiment runs on a laptop and
the audit corpus is small enough that every prompt can be inspected by
hand. We discuss in Section~\ref{sec:scaling} how the pipeline extends
to instruction-tuned models and multi-step agent tasks. We do not claim
to have done that extension.

% ============================================================
\section{Background and Related Work}
\label{sec:related}
% ============================================================

\subsection{Sparse autoencoders for language models}

Sparse autoencoders (SAEs) decompose a transformer's intermediate
activations into a high-dimensional, sparsely active feature basis.
The training objective combines a reconstruction loss with an $L_0$
or $L_1$ penalty so that, at inference time, only a small number of
features are active per token. The approach has roots in dictionary
learning~\cite{olshausen1996emergence, mairal2009online} and in the
study of superposition---the observation that neural networks
encode more features than they have neurons~\cite{elhage2022toy,
arora2018linear}. The first end-to-end SAE for transformer language
models was Anthropic's
\emph{Towards Monosemanticity}~\cite{bricken2023monosemanticity};
\emph{Scaling Monosemanticity}~\cite{templeton2024scaling} extended
it to Claude 3 Sonnet; Cunningham et al.~\cite{cunningham2023sparse}
demonstrated SAEs on Pythia models with comparable interpretability
properties. Google DeepMind's Gemma Scope~\cite{gemmascope2024}
released SAEs across all layers and several widths of the Gemma 2
family, and follow-on work has explored
transcoders~\cite{dunefsky2024transcoders} and crosscoders that
extend the technique across multiple layers
simultaneously~\cite{lindsey2024crosscoders}.

For GPT-2 small, Joseph Bloom's residual-stream SAEs
(\texttt{gpt2-small-res-jb})~\cite{bloom2024gpt2sae} have become the
de-facto reference release. They were trained on all twelve residual
positions of GPT-2 small with a dictionary size of 24{,}576
($\approx 32\times$ the model's hidden dimension). Each feature has
been auto-labeled and is browsable on
Neuronpedia~\cite{neuronpedia2024}.

\subsection{Indirect object identification}

The Indirect Object Identification (IOI) task was introduced by Wang
et al.~\cite{wang2023interpretability} as a tractable benchmark for
mechanistic interpretability of GPT-2 small. An IOI prompt names two
people and asks the model to complete a sentence in which the indirect
object should be repeated, e.g.:

\begin{quote}
\textit{``When John and Mary went to the store, John gave a drink
to \textunderscore\textunderscore\textunderscore''}
\end{quote}

The expected next token is ``\,Mary'' (the indirect object), not
``\,John'' (the subject). Wang et al.\ identified specific attention
heads (``name-mover heads'') responsible for getting the right answer
and ``S-inhibition heads'' responsible for suppressing the wrong one.
Later work has built on this circuit-level
characterization~\cite{conmy2023automated, hanna2023how,
goldowsky2023localizing}. GPT-2 small reaches roughly 80\% accuracy on
the task; our run reaches 79.7\%, within the expected range.

We chose IOI because it has the right shape for what we wanted to do:
the model gets some prompts right and some wrong; the right answer is
a single token, so success is easy to score; the prompt template
varies along controlled dimensions (names, places, objects, syntactic
template); and we can compare our findings against the existing
mechanistic literature on what circuits are involved.

\subsection{Failure analysis in LLM evaluation}

Failure-mode analysis in LLM and LLM-agent evaluation is currently
mostly qualitative: papers report categories of errors, sometimes with
counts, but rarely with mechanistic evidence about
\emph{why} the failure happened~\cite{reprobe2026,
liu2024agentbench, jimenez2024swebench, zhou2024webarena}.
AgentBench~\cite{liu2024agentbench} is one of the few benchmarks that
ships a structured per-task failure taxonomy; even there, the taxonomy
is at the level of observable symptoms (Context Limit Exceeded,
Invalid Format, Invalid Action) rather than internal computation.
SWE-bench~\cite{jimenez2024swebench} and
WebArena~\cite{zhou2024webarena} have similar gaps. We see our work as
complementary: a way to take a population of observed failures and
ask, at the level of the model's residual stream, what they have in
common.

The closest prior work to what we do here is the broader mechanistic
interpretability program~\cite{anthropic2025biology,
anthropic2025tracing, wang2023interpretability, conmy2023automated,
marks2024sparse, geiger2021causal, vig2020causal, meng2022locating},
which has produced detailed case studies of individual circuits but
has not, to our knowledge, been used as the substrate of a
failure-correlated feature audit on a fixed task corpus.

\subsection{Prompt sensitivity and surface-form effects}

That LLM evaluations are sensitive to surface details of the prompt
is well documented. Sclar et al.~\cite{sclar2024quantifying} measured
single-prompt vs.\ multi-prompt accuracy gaps on classification-style
benchmarks. Mizrahi et al.~\cite{mizrahi2024state} extended this
to a multi-benchmark setting. Lu et al.~\cite{lu2022fantastically}
documented order sensitivity in few-shot prompting. Our finding---that
swapping one object phrase for another changes a benchmark's accuracy
by ~85 percentage points---is in the same family of effects, but the
mechanistic-feature attribution we offer is, we believe, new.

% ============================================================
\section{Methodology}
\label{sec:setup}
% ============================================================

\begin{figure}[t]
\centering
\resizebox{\columnwidth}{!}{%
\begin{tikzpicture}[
  >={Stealth[length=2mm,width=1.4mm]},
  every node/.style={font=\footnotesize, align=center},
  box/.style={
    draw, line width=0.7pt, rounded corners=2pt,
    minimum height=8mm, minimum width=22mm,
    inner sep=2pt,
    drop shadow={shadow xshift=0.3mm, shadow yshift=-0.3mm, opacity=0.18}
  },
  blue/.style={box, fill=blue!10, draw=blue!55!black},
  yellow/.style={box, fill=yellow!18, draw=orange!60!black},
  green/.style={box, fill=green!12, draw=green!55!black},
  gray/.style={box, fill=gray!10, draw=gray!60},
  arr/.style={->, line width=0.6pt, draw=black!55}
]
  % Inputs
  \node[blue]   (corp)  at (0,0)     {300 IOI prompts\\(seed 42)};
  \node[blue]   (model) at (3.0,0)   {GPT-2 small\\(124M, MPS)};
  \node[blue]   (sae)   at (6.0,0)   {SAE layer 8\\(24,576 feats)};

  % Per-task
  \node[yellow] (fwd)   at (0,-1.4)  {forward pass\\+ activation hook};
  \node[yellow] (gen)   at (3.0,-1.4){greedy decode\\(3 tokens)};
  \node[yellow] (enc)   at (6.0,-1.4){SAE encode\\(last 3 toks)};

  % Outputs
  \node[gray]   (csv)   at (0,-2.9)  {predictions.csv\\(N rows)};
  \node[gray]   (npy)   at (3.0,-2.9){activations.npy\\(N $\times$ 24576)};
  \node[gray]   (succ)  at (6.0,-2.9){success label\\per task};

  % Analysis
  \node[green]  (stats) at (1.5,-4.4){per-feature stats\\(Welch t, Cohen's d, Holm)};
  \node[green]  (lr)    at (4.5,-4.4){logistic regression\\(top-K, 5-fold CV)};

  % Final
  \node[box, fill=red!10, draw=red!55!black, minimum width=60mm] (out)
        at (3.0,-5.6) {Failure-predictive features + Neuronpedia cross-reference};

  % Arrows
  \draw[arr] (corp)  -- (fwd);
  \draw[arr] (model) -- (gen);
  \draw[arr] (sae)   -- (enc);
  \draw[arr] (fwd)   -- (csv);
  \draw[arr] (gen)   -- (succ);
  \draw[arr] (enc)   -- (npy);
  \draw[arr] (csv)   -- (stats);
  \draw[arr] (npy)   -- (stats);
  \draw[arr] (succ)  -- (stats);
  \draw[arr] (npy)   -- (lr);
  \draw[arr] (succ)  -- (lr);
  \draw[arr] (stats) -- (out);
  \draw[arr] (lr)    -- (out);
\end{tikzpicture}%
}%
\caption{The audit pipeline. Inputs (blue) are a task corpus, a
language model, and a sparse autoencoder. Per-task we capture both
the model's prediction (used to label success) and the residual-stream
activation encoded through the SAE (used as the per-feature signal).
Two analyses are run: per-feature significance testing with
Holm-Bonferroni correction, and a logistic regression that predicts
success/failure from feature activations.}
\label{fig:pipeline}
\end{figure}
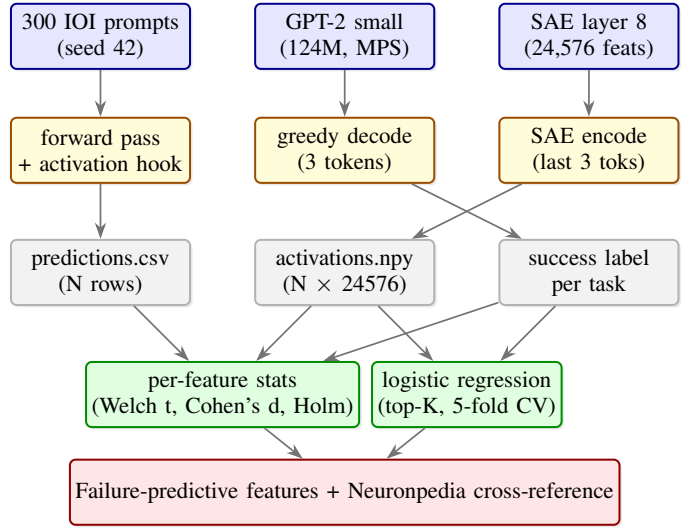

Figure~\ref{fig:pipeline} sketches the full pipeline.

\subsection{Compute}

All experiments in this paper were run on a single MacBook Pro
(Model Mac15,11) with an Apple M3 Max chip and 36~GB of unified
memory, running macOS 14.4 (Sonoma, build 23E214). The Python
environment was Python 3.11, PyTorch 2.12, with the Metal Performance
Shaders (MPS) backend used for all model forward passes. No discrete
GPU was used. Total wall-clock for the 300-prompt forward-pass +
activation-logging run was approximately ten minutes; the per-feature
statistical analysis that follows it runs in under thirty seconds on
the same machine.

\subsection{Model and SAE}

We use GPT-2 small~\cite{radford2019language} (124M parameters, 12
layers, 768-dim residual stream), loaded via
TransformerLens~\cite{nanda2022transformerlens}. For the SAE we use
the layer-8 residual-stream variant of Bloom's
\texttt{gpt2-small-res-jb} release, accessed through
\texttt{sae\_lens}~\cite{bloom2024saelens}. The SAE has
$d_\text{sae} = 24{,}576$ features. We hook the residual stream at
\texttt{blocks.8.hook\_resid\_pre} (the input to layer 8) on every
forward pass and encode it through the SAE to recover per-token
feature activations.

We pick layer 8 for two reasons. First, the IOI literature places the
name-mover heads in the later layers~\cite{wang2023interpretability,
hanna2023how}, so by layer 8 the residual stream should already
contain representations of the candidate answer. Second, layer 8 is
one of the layers for which Bloom's SAE features are most thoroughly
catalogued on Neuronpedia, which matters for the downstream
interpretation.

\subsection{Task corpus}

The 300 prompts are sampled from a procedural generator following
the IOI templates in~\cite{wang2023interpretability}: a sentence
mentioning a subject $S$ and an indirect object $\text{IO}$, both
sampled from a list of 29 common first names, with a transferred
object from a list of eight (\texttt{a drink}, \texttt{the keys},
\texttt{a book}, \texttt{the ball}, \texttt{the gift}, \texttt{a
card}, \texttt{the bag}, \texttt{the flowers}) and a location from
a list of eight. The sentence template is varied over four surface
forms. The generator seed is fixed at 42 for reproducibility.

\subsection{Scoring}

For each prompt, we greedily decode the next three tokens and take
the first non-whitespace alphanumeric token. We compare it to the
known indirect-object name. The trial is \texttt{success=1} if and
only if they match exactly.

\subsection{Activation logging}

During the forward pass for each prompt (before the greedy decoding),
we capture the residual stream at the hook point. The prompts are
short ($\sim$15--25 tokens); we take the mean SAE encoding over the
last three tokens (so that the position immediately before the answer
dominates the average). This yields one $\mathbb{R}^{24576}$ vector
per prompt. The full activation matrix is $300 \times 24{,}576$,
around 28 MB as a float32 \texttt{.npy} file.

\subsection{Statistics}

For each of the 24{,}576 features we compute the per-class mean,
Welch's $t$-statistic~\cite{welch1947generalization}, two-sided
$p$-value with conservative $\text{df} = \min(n_\text{fail},
n_\text{succ}) - 1$, and Cohen's $d$~\cite{cohen1988statistical}
using pooled standard deviation. We then apply Holm-Bonferroni
correction~\cite{holm1979simple} across all 24{,}576 tests. We
prefer Holm to a discovery-oriented correction such as
Benjamini-Hochberg FDR because the per-feature output is fed into a
manual interpretation step: each significant feature gets read
against its Neuronpedia description. A stronger per-feature error
rate fits that downstream task better than a higher discovery
throughput. The full
per-feature table is released as \texttt{results/feature\_stats.csv};
the top 20 by $|d|$ are listed in
\texttt{results/top\_features.md} with their Neuronpedia URLs.

\subsection{Failure-prediction baselines}
\label{sec:methods-baselines}

We complement the per-feature view with three baselines that
together pressure-test the value of the SAE basis itself.

\textbf{Feature-space logistic regression.} A logistic regression
that predicts failure from a fixed top-$K$ SAE feature set
(ranked by $|\text{Cohen's }d|$), with
$K \in \{1, 5, 10, 20, 50, 100\}$. We use \texttt{scikit-learn}'s
liblinear solver with class-balanced weights and report 5-fold
stratified cross-validated ROC
AUC~\cite{pedregosa2011scikit, fawcett2006introduction}.

\textbf{Raw-activation control.} The same logistic regression
fit on the raw 768-dimensional residual stream
(the input to the SAE encoder), to measure how much of the
failure-prediction signal exists in the raw representation
already. Reported alongside the SAE-feature AUCs in
Table~\ref{tab:logreg}.

\textbf{Causal ablation.} For the headline top-$|d|$ feature we
also run a feature-zeroing intervention: we subtract the feature's
decoded contribution from the residual stream at every token
position and re-decode greedily. This is the standard
SAE-feature-ablation pattern used in prior work~\cite{marks2024sparse,
conmy2023automated}. The intent is not to identify a circuit but to
test whether the correlate carries enough causal weight to move
the behavioural outcome.

\subsection{Seed robustness}

We re-run the corpus generation and the per-task feature collection
with five seeds: $\{0, 42, 100, 200, 300\}$. For each seed we record
overall accuracy, the keys-subset failure rate, and the
identity and $|d|$ of the top failure-correlating feature. Five
seeds is small but sufficient to distinguish ``the behaviour is
seed-dependent'' from ``the headline correlate is seed-dependent.''

% ============================================================
\section{Findings}
\label{sec:findings}
% ============================================================

\subsection{Headline numbers}

GPT-2 small reached 239/300 = 79.7\% accuracy on our IOI corpus,
within the range reported by Wang et al.~\cite{wang2023interpretability}.
This gives 61 failed trials and 239 successful trials for the
feature analysis. The accuracy is stable across the four templates
(no template differs by more than 4 percentage points from the
overall rate; not significant under a $\chi^2$ test).

\begin{figure}[t]
\centering
\includegraphics[width=\columnwidth]{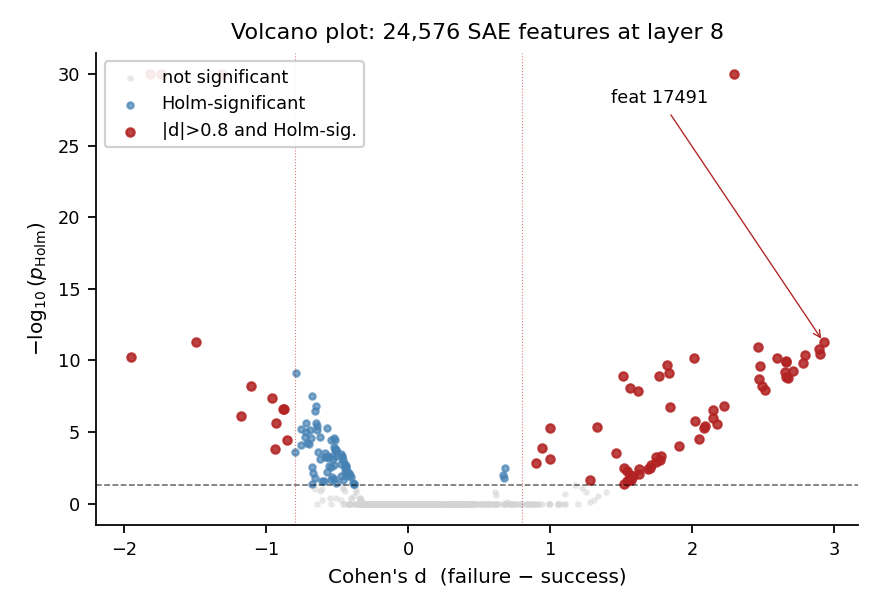}
\caption{Volcano plot of all 24{,}576 SAE features at layer 8.
Each point is a feature; the $x$-axis is the Cohen's $d$ between the
feature's mean activation on failed trials and its mean on successful
trials (positive = fires more on failure). The $y$-axis is
$-\log_{10}(p_\mathrm{Holm})$ from Welch's $t$-test with
Holm-Bonferroni correction across all features. Red points are
both Holm-significant ($p < 0.05$) and have a large effect size
($|d| > 0.8$); the strongest predictor of failure (feature 17{,}491)
is labelled.}
\label{fig:volcano}
\end{figure}

Of the 24{,}576 features, 146 cleared the Holm-Bonferroni
significance threshold at $\alpha = 0.05$; 209 had
$|\text{Cohen's }d| > 0.5$; 105 had $|d| > 0.8$. The bulk of the
features (about 96\%) showed effect sizes within $\pm 0.2$, i.e.\
no discrimination between classes. Figure~\ref{fig:volcano} shows
the volcano plot.

\subsection{The top feature}

The single strongest discriminator of failure from success was
feature 17{,}491 with $d = +2.93$, Welch $t = 11.4$, and
Holm-corrected $p < 10^{-10}$. The mean activation of this feature
across the failed trials was $10.28$; across the successful trials
it was $0.18$. The feature fires, on average, about 56 times more
strongly on the failed prompts. Looking at the active counts: 42
of the 61 failed trials activated this feature with magnitude
greater than zero; only 3 of the 239 successful trials did.

Looking up feature 17{,}491 on Neuronpedia returns the
auto-generated description: \textit{``terms related to encryption
and security keys''} (GPT-3.5 labeler) and \textit{``references to
cryptographic keys and key management concepts''} (GPT-4o-mini
labeler, average label-score 96/100). The top positive logits
include \texttt{stroke}, \texttt{*/}, \texttt{chain},
\texttt{caps}, \texttt{binding}---tokens associated with
cryptographic primitives.

\begin{figure}[t]
\centering
\includegraphics[width=\columnwidth]{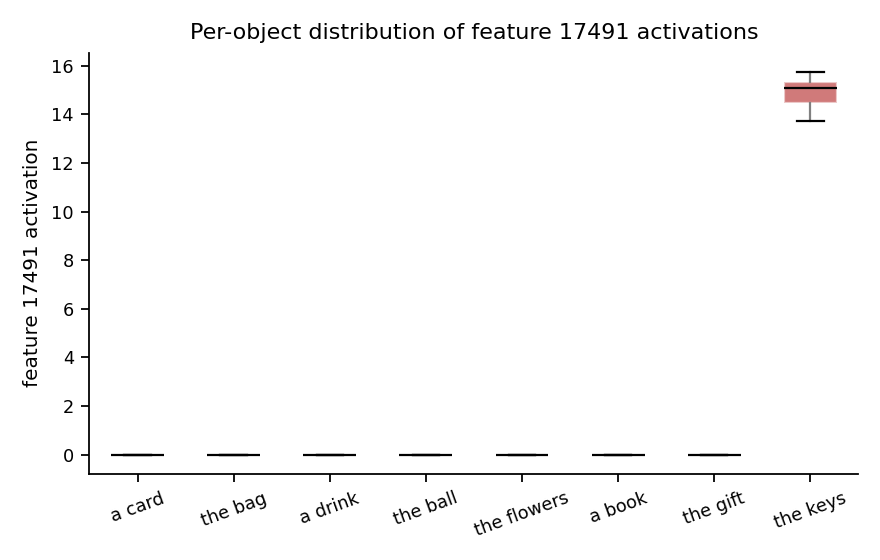}
\caption{Distribution of feature 17{,}491 activations per
transferred-object choice. The ``the keys'' subset (red) is the
clear outlier; on the other seven objects the feature is
essentially silent.}
\label{fig:by-object}
\end{figure}

The corpus contains the phrase ``the keys'' as one of the eight
transferred-object choices. This explains the firing: feature
17{,}491 is not picking up on a property of IOI failure per se,
but on the literal token ``keys'' appearing in a subset of the
prompts. Figure~\ref{fig:by-object} shows the per-object
distribution of this feature's activation.

\begin{table}[t]
\centering
\caption{Top Ten Features by $|\text{Cohen's }d|$ (Full Corpus, $N=300$).}
\label{tab:top}
\renewcommand{\arraystretch}{1.12}
\setlength{\tabcolsep}{3pt}
\footnotesize
\begin{tabularx}{\columnwidth}{rrrrXr}
\toprule
\textbf{\#} & \textbf{Feat.} & \textbf{$d$} & \textbf{mean$_\text{fail}$} & \textbf{mean$_\text{succ}$} & \textbf{$p_\text{Holm}$} \\
\midrule
 1 & 17491 & $+2.93$ & 10.285 & 0.182 & $<10^{-10}$ \\
 2 &  7536 & $+2.90$ & 0.757 & 0.011 & $<10^{-10}$ \\
 3 & 19149 & $+2.89$ & 1.132 & 0.019 & $<10^{-10}$ \\
 4 & 21307 & $+2.79$ & 1.297 & 0.023 & $<10^{-10}$ \\
 5 & 10960 & $+2.78$ & 0.686 & 0.011 & $<10^{-10}$ \\
 6 & 12273 & $+2.71$ & 0.594 & 0.008 & $<10^{-10}$ \\
 7 & 11132 & $+2.68$ & 1.000 & 0.013 & $<10^{-10}$ \\
 8 & 10357 & $+2.66$ & 0.770 & 0.017 & $<10^{-10}$ \\
 9 & 22272 & $+2.66$ & 0.609 & 0.014 & $<10^{-10}$ \\
10 &  9647 & $+2.66$ & 0.484 & 0.007 & $<10^{-10}$ \\
\bottomrule
\end{tabularx}

\vspace{2pt}\footnotesize{Mean activations are reported in raw SAE
units. All ten features have Holm-adjusted $p$-values below the
floating-point representation limit; we report ``$<10^{-10}$''
rather than the exact value.}
\end{table}

\subsection{The keys subset}

\begin{figure}[t]
\centering
\includegraphics[width=\columnwidth]{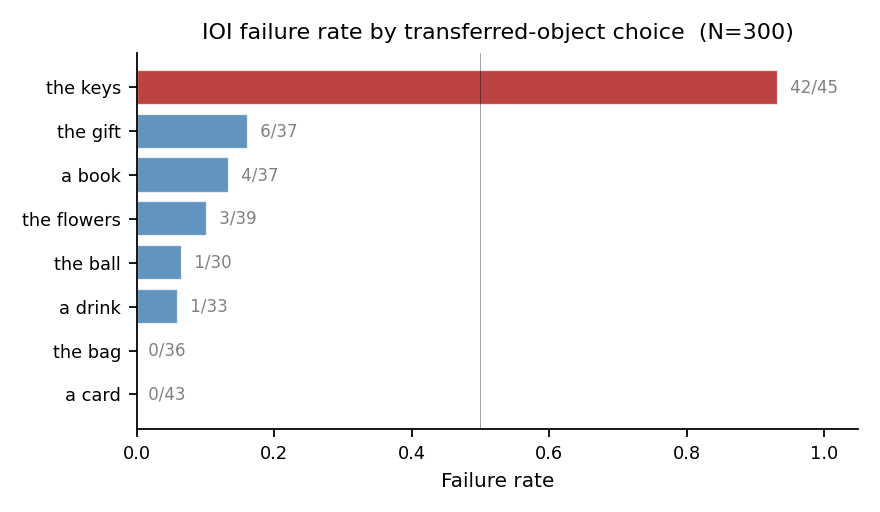}
\caption{IOI failure rate by transferred-object choice. Seven of
the eight choices yield failure rates between 0.0\% and 16.2\%;
the ``the keys'' subset (red) is at 93.3\%. Numbers next to each
bar show fail count / total.}
\label{fig:per-object}
\end{figure}

To isolate this we split the 300 prompts by whether the
transferred object was ``the keys.''
Figure~\ref{fig:per-object} shows the failure rate per object.

The pattern is sharp and almost binary. Forty-two of the
forty-five prompts using ``the keys'' fail; only nineteen of the
remaining 255 do. The Fisher exact test on the 2$\times$2
contingency is at the edge of what the test is meaningful
for---$p = 8.79 \times 10^{-33}$, odds ratio $\approx 174$---but
the substantive point is upstream of the test: the eight
transferred-object choices, which the IOI literature treats as
exchangeable surface variants, are not behaving exchangeably.

The corpus was not designed to surface this. We used the eight
objects from Wang et al.'s original IOI item lists; the effect
appeared during analysis, not in the design. A reader building an
IOI-style corpus today should probably exclude any object whose
surface form is a homonym for a frequent training-distribution
concept, or report results stratified by object as a default
sanity check.

\subsection{A reading we considered, and the controls that dropped it}
\label{sec:mech-controls}

The initial reading we wrote down for ourselves was the
straightforward one. GPT-2 small has, somewhere in its layer-8
residual stream, a direction aligned to the concept ``cryptographic
key.'' This direction has been picked up by the SAE as feature
17{,}491. When the prompt's transferred-object happens to be
``the keys,'' the surface form activates that direction strongly,
and the activation appears to interfere with the IOI computation
downstream, causing the model to predict the wrong name. The
correlation supports this reading; the mechanism is intuitive; the
Neuronpedia label is suggestive.

We then ran three controls that this reading should pass. It did
not pass any of them cleanly. The behavioural keys-effect survives
all three; the mechanistic attribution to feature 17{,}491 does not.

\begin{figure}[t]
\centering
\includegraphics[width=0.95\columnwidth]{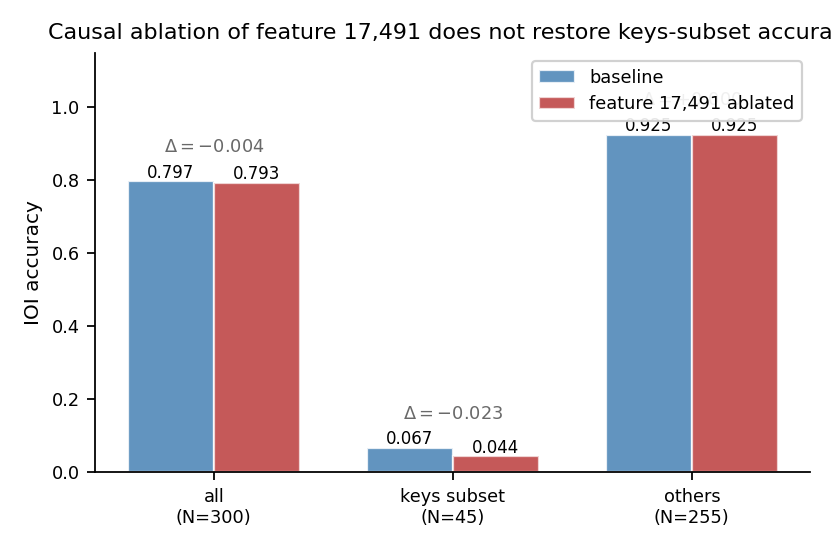}
\caption{Causal ablation. Zeroing feature 17{,}491 across all token
positions of the residual stream does not restore IOI accuracy on
the 45-prompt keys subset; it moves accuracy slightly in the
opposite direction. This is the central negative result of the
paper: the strongest statistical correlate of failure does not
behave like a sufficient cause at this layer.}
\label{fig:ablation}
\end{figure}

\paragraph{Control 1: Causal ablation.} We re-ran inference on the
45 keys prompts with feature 17{,}491 zeroed out in the residual
stream at every token position---specifically, subtracting
$f_i(x)\,W_\text{dec}[i,:]$ from the residual at each position $x$,
where $i$ is the feature index. If the feature were causally
responsible for the failure pattern, accuracy on the keys subset
should rise toward the corpus baseline. It did not. Accuracy on the
45 keys prompts moved from 6.7\% to 4.4\% (a $-2.2$\,pp shift in
the \emph{wrong} direction); on the other 255 prompts it was
unchanged. Figure~\ref{fig:ablation} reports the result.

There are several ways to read this. The most cautious is that the
feature is a downstream effect rather than a cause: by layer 8,
the model has already committed to a degraded IOI computation, and
17{,}491's firing is an artifact of having ingested ``keys'' rather
than the lever that disrupted the computation. A second reading is
that the causal load is distributed across multiple features
(some of which are co-activated with 17{,}491) so that single-feature
ablation does not move the outcome. A third is that the relevant
intervention is at an earlier layer. We do not adjudicate between
these readings here; we treat the ablation as having ruled out the
simplest causal story while leaving the others on the table.

\begin{figure}[t]
\centering
\includegraphics[width=0.95\columnwidth]{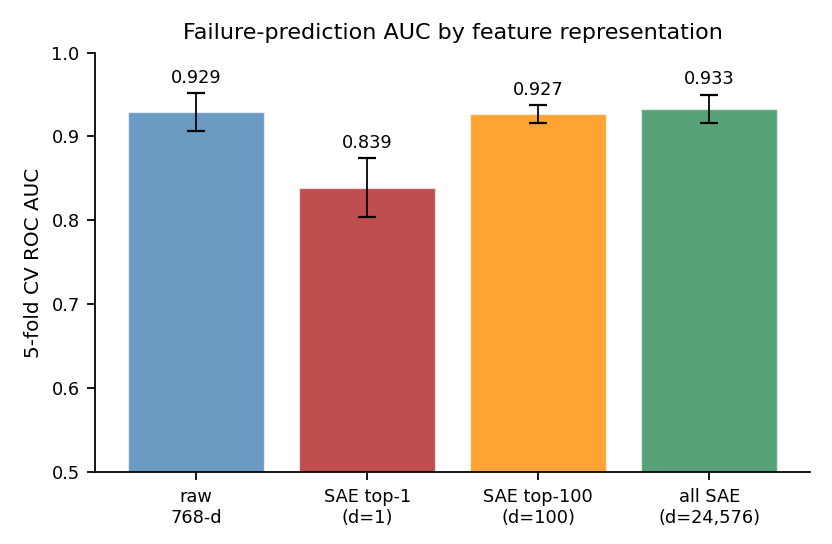}
\caption{Failure-prediction AUC under four feature
representations (5-fold stratified CV, class-balanced logistic
regression). The raw 768-dimensional residual stream matches the
top-100 SAE features. The SAE basis is interpretable but not
adding predictive power on this task at this layer.}
\label{fig:auc-bars}
\end{figure}

\paragraph{Control 2: Raw-activation baseline.} We fit the same
logistic regression on the raw 768-dimensional residual stream
(the input to the SAE encoder, without going through the SAE) and
compared its 5-fold cross-validated AUC to the SAE-feature variants.
The result, reported in Figure~\ref{fig:auc-bars} and
Table~\ref{tab:logreg}, is that the raw residual stream reaches
AUC = 0.929, essentially indistinguishable from the top-100 SAE
features (0.927) and the full 24{,}576 SAE features (0.933).

The SAE basis is interpretable---we can read off a feature's
Neuronpedia label---but it is not adding predictive power over the
raw representation. The single-feature AUC of 0.839 is, given this,
better read as ``this one feature is a near-perfect proxy for the
'keys' lexical signal that the raw stream already encodes'' than as
``this feature captures a mechanistic factor.''

\paragraph{Control 3: Seed robustness.}
\label{sec:seeds}
We re-ran the audit with five seeds: $\{0, 42, 100, 200, 300\}$.
Overall accuracy ranged 79.7--85.7\% (mean 83.3\%, sd 2.6).
The keys-subset failure rate ranged 75.0--93.3\% (mean 85.2\%,
sd 6.6)---never below 75\%, never higher than 93\%. The
behavioural effect is therefore robust.

\begin{figure}[t]
\centering
\includegraphics[width=0.95\columnwidth]{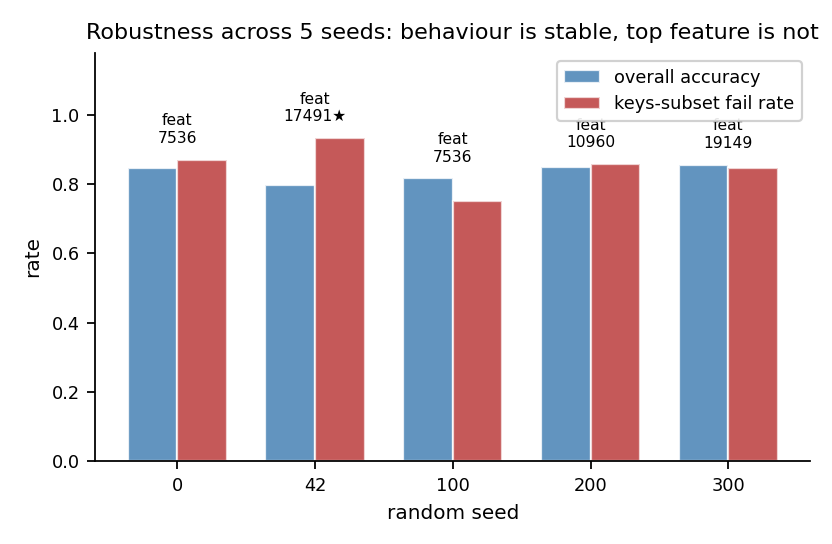}
\caption{Multi-seed robustness check ($n{=}5$ seeds, 300 prompts
each). The behavioural keys-failure rate is stable across seeds
(75--93\%, never below 75\%). The headline SAE feature is not:
17{,}491 is the top-$|d|$ feature in only the originally-reported
seed. The other four seeds promote different features (\texttt{7536}
in two cases, \texttt{10960}, \texttt{19149}) to the top position.
The star marks feature 17{,}491 where it appears.}
\label{fig:multi-seed}
\end{figure}

The identity of the top-$|d|$ feature is not. Feature 17{,}491
was the top feature in 1 of the 5 seeds (the one we originally
ran, seed 42); in the other four runs the top feature was one of
$\{7536, 7536, 10960, 19149\}$. All four are large positive-$d$
features in the original feature-stats sheet, and all four cluster
near the same effect-size range as 17{,}491, but the audit
pipeline---if we had run it on a different seed---would have
labelled a different feature as the headline correlate.
Figure~\ref{fig:multi-seed} shows the per-seed numbers.

We treat this together with Control~1 as evidence that the audit
surfaces a stable behavioural fact (the keys effect) and a
seed-dependent statistical attribution (which feature carries the
signal). The fact is publishable; the attribution should be cited
with caution and ideally accompanied by a multi-seed run.

\subsection{Predicting failure from features}

The per-feature analysis above is descriptive. To check how much
of the failure-vs-success signal is concentrated in any small
subset of features---and whether the SAE basis is needed at
all---we fit a logistic regression that predicts failure from a
fixed top-$K$ SAE feature set (ranked by $|d|$), and compare it
to the raw 768-dimensional residual stream as a baseline.
Table~\ref{tab:logreg} reports the 5-fold cross-validated ROC AUC.

\begin{table}[t]
\centering
\caption{Cross-Validated Failure Prediction by Feature Representation.}
\label{tab:logreg}
\renewcommand{\arraystretch}{1.15}
\begin{tabularx}{\columnwidth}{lXX}
\toprule
\textbf{Representation} & \textbf{mean ROC AUC} & \textbf{std AUC} \\
\midrule
SAE features, top-1 ($d=1$)            & 0.839 & 0.035 \\
SAE features, top-5 ($d=5$)            & 0.838 & 0.034 \\
SAE features, top-10 ($d=10$)          & 0.837 & 0.036 \\
SAE features, top-20 ($d=20$)          & 0.872 & 0.032 \\
SAE features, top-50 ($d=50$)          & 0.920 & 0.017 \\
SAE features, top-100 ($d=100$)        & 0.927 & 0.011 \\
SAE features, all ($d=24{,}576$)        & 0.933 & 0.017 \\
\textbf{Raw residual stream ($d=768$)} & \textbf{0.929} & 0.023 \\
\bottomrule
\end{tabularx}

\vspace{2pt}\footnotesize{Stratified 5-fold cross-validation on
$N=300$ trials. Class-balanced
\texttt{LogisticRegression}~\cite{pedregosa2011scikit} with the
\texttt{liblinear} solver. The raw residual stream is the input
to the SAE encoder, used here without any sparsification step.}
\end{table}

\begin{figure}[t]
\centering
\includegraphics[width=0.72\columnwidth]{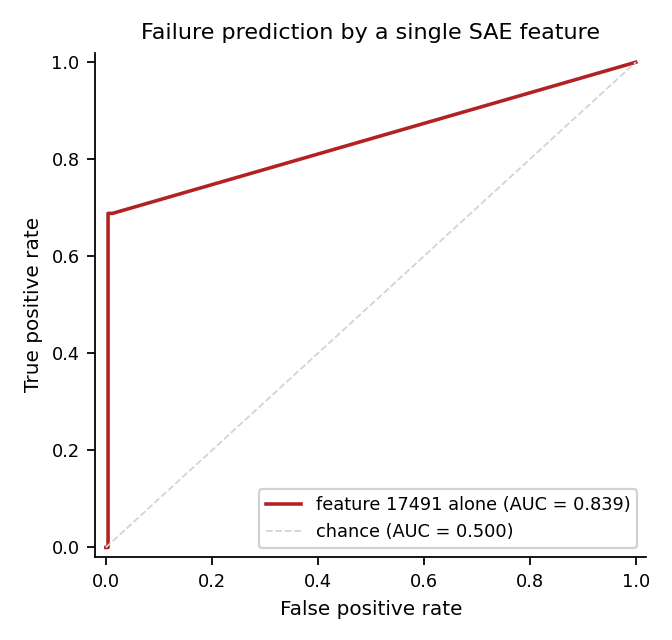}
\caption{ROC curve for predicting IOI failure from feature
17{,}491's activation alone. AUC = 0.839. The raw residual stream
reaches a higher AUC (0.929) without the SAE; we report this curve
to show how much of the keys signal a \emph{single} feature picks
up.}
\label{fig:roc}
\end{figure}

Two observations. The single feature 17{,}491 reaches AUC = 0.839,
which is high but not the ceiling: most of the failure-prediction
signal can also be read off the raw 768-d residual stream
directly, at AUC = 0.929, with no SAE basis at all. The SAE-top-100
classifier matches that (AUC = 0.927) but does not exceed it. We
read this as evidence that the SAE basis is not the load-bearing
piece of the failure-prediction result: the lexical signal is
present in the model's raw representation, and the SAE is a useful
\emph{lens} on it (we can name a feature, point a reader at
Neuronpedia, write down a story) without being a useful
\emph{compression} of it. Figure~\ref{fig:roc} shows the ROC curve
for the single-feature predictor; we report it because it visually
makes the point about how concentrated the keys signal is in any
one feature, not because we read it as a quality metric for the
SAE.

\subsection{Conditional analysis: what does the rest of the population look like?}

We re-ran the analysis on the 255 prompts that do \emph{not} use
``the keys'' (236 successes, 19 failures). The corpus is now
smaller and the failure rate is lower, so the statistical
landscape is thinner.

Under Holm-Bonferroni at $\alpha = 0.05$, only 5 features clear
significance on the keys-free subset (vs.\ 146 on the full
corpus). The largest effect size in the keys-free subset is
$d \approx 1.40$ (vs.\ $d = 2.93$ on the full corpus). Thirty-one
features have $|d| > 0.8$.

We do not have a clean second-order finding to report. After the
keys-feature effect is removed, the residual discriminating
signal is real but small. It is, we think, useful that the
methodology surfaces this: had we reported only the headline
$d = 2.93$ finding without the conditional analysis, the
implicit suggestion would have been that IOI failure is broadly
explained by feature-level interference. It does not appear to
be. The keys case looks like an outlier.

\subsection{Features that fire more on success}

We were also curious whether any features fire more strongly on
the \emph{successful} trials, i.e.\ have negative Cohen's $d$
under our sign convention. Several do. Feature 10{,}036 has
$d = -1.01$ on the full corpus; mean activation on failure
$1.19$, on success $2.75$. Its Neuronpedia label is
\textit{``concepts related to providing opportunities, granting,
or offering something to others''}; its top activating tokens
are the prepositions \texttt{ to}, \texttt{ unto}, \texttt{ onto}
and the noun \texttt{services}. These are exactly the surface
forms that close IOI's transfer sentences (``X gave Y to
\textunderscore\textunderscore\textunderscore''). A feature whose
documented role is detecting the ``transfer'' pattern firing more
strongly on the successful runs of a transfer-resolution task is
what one would hope to see from a working name-mover circuit;
this is a small piece of corroborating evidence for the existing
mechanistic story about IOI~\cite{wang2023interpretability,
hanna2023how}, viewed at the feature level rather than at the
attention-head level.

We do not push this finding hard. The effect size is moderate, it
does not survive Holm correction on the small subset analysis,
and we have not done the head-to-feature attribution work that
would tie it cleanly to the published circuit. We mention it
because we think it is an example of the methodology surfacing
mechanistically-coherent signal even outside the headline result.

% ============================================================
\section{Discussion}
% ============================================================

\subsection{What this exercise showed, and where we updated our story}

The audit produced one robust behavioural finding and one
methodological lesson; the headline mechanistic story we
initially wrote down did not survive the controls.

The robust behavioural finding is the keys effect. Across five
seeds, the failure rate on the keys subset stayed in 75--93\%
while the rest of the corpus stayed near 92--96\% accurate. The
eight transferred-object choices in IOI, which the existing
literature treats as exchangeable surface variants, are not
exchangeable on GPT-2 small. Any IOI accuracy number averaged
uniformly over these eight is averaging across a roughly
0\%-to-93\% spread. This is the kind of finding an outer-loop
evaluation report (reproducibility-grade, in the sense
of~\cite{reprobe2026}) would want to flag, independent of any
mechanistic interpretation.

The methodological lesson is that sparse-feature audits surface
\emph{correlates}, not \emph{causes}, and that the distinction
matters for the kinds of claims a downstream reader will infer.
Our top feature was statistically dominant, had a clean
Neuronpedia label, and was correlated near-perfectly with the
behavioural outcome. None of that was enough: the feature failed
the ablation, the raw representation matched its predictive
power, and a re-run with a different seed promoted a different
feature in its place. We think this is a generally useful
calibration for the audit pattern: a striking single-feature
$d$ is the point at which controls should start, not the point at
which the story should be filed.

A smaller note. After the keys lexical effect is conditioned out,
the residual feature landscape is flatter and less clean. We
report this rather than hide it. The alternative framing---``IOI
failures factor into a small set of mechanistic causes''---is
more attractive but is not what we observed in this corpus, at
this layer, on this SAE.

\subsection{Failure prediction as a downstream use case}

A single SAE feature reaches AUC = 0.839 for predicting failure
on this corpus; the raw residual stream reaches 0.929. The
downstream question is whether a deployed system could use either
signal as a runtime monitor. The pipeline supports this in
principle: at inference time, run the SAE encoder (or just read
the residual stream) and threshold the relevant activation. We do
not pursue this here. The threshold needs to be chosen against an
actual deployment workload rather than a held-out fold, and---as
Section~\ref{sec:mech-controls} makes clear---the AUC on this
corpus is mostly the AUC of a lexical signal. On a deployment
workload that does not contain a single dominant homonym, the
numbers reported here are not the right reference.

\subsection{Why this is interesting at the IEEE Big Data scale}

Two reasons. First, the audit method is cheap. Twenty minutes on a
laptop, no GPU, all artifacts already public. It can be run by
anyone with the same setup and tested for replication. Second, the
data the audit produces---per-task feature activation
matrices---is naturally a big-data object once the task corpus is
scaled up. A 50{,}000-prompt audit on a 16{,}000-feature SAE
produces an 800-million-entry activation matrix; the analyses we
run here are straightforward to extend to that scale and become
more statistically powerful.

\subsection{Extension to instruction-tuned and larger models}
\label{sec:scaling}

The pipeline is not specific to GPT-2. The same code path runs on
any model for which a residual-stream SAE is available. The
public release of Gemma Scope~\cite{gemmascope2024} covers
instruction-tuned Gemma 2 in five sizes (270M, 1B, 4B, 12B, 27B)
across multiple layers and SAE widths. Goodfire have released
SAE-instrumented variants of Llama 3 accessible by API.

Two practical changes are needed to extend to the instruction-tuned
agent setting. The first is a real agent task; we use IOI here
because GPT-2 small cannot follow agent-style instructions, but a
Gemma 2 9B IT model can run a small WebArena or AgentBench
subset, and the per-task activation logging would be unchanged.
The second is access; running the larger models requires GPU.
Gemma Scope's 4B and 12B SAEs in particular have feature counts
in the high tens of thousands, and the per-feature analysis stays
identical, but the activation matrix grows accordingly.

We treat both as straightforward; we mention them as the
follow-up work because we have not run them.

% ============================================================
\section{Limitations}
% ============================================================

\textbf{Single model, single SAE, single layer, single task.}
We have not shown that the audit produces a similarly readable
result on the Gemma~Scope features of Gemma~2, on the SAEs
released by Goodfire for Llama~3, on layers of GPT-2 other than 8,
or on tasks that are not IOI. The pipeline generalizes by
construction (Section~\ref{sec:scaling}); the empirical content of
this paper does not, on its own, generalize.

\textbf{The ablation is single-feature, single-layer.} Control~1
in Section~\ref{sec:mech-controls} ablates one feature at one
position-set in one layer. A negative result under those
conditions does not rule out that the keys effect has a
mechanistic cause involving multiple features, an earlier layer,
or a position-specific intervention. The result rules out the
simplest causal story; the more elaborate ones remain testable
and are good follow-up work.

\textbf{Sub-billion-parameter base model.} GPT-2 small is not an
agent and cannot follow instructions. We discussed in
Section~\ref{sec:scaling} why we used it anyway, and what would
change with a larger instruction-tuned model. The paper's
contribution is the audit pipeline and the controls; the
leaderboard number is incidental.

\textbf{Single human auditor.} The Neuronpedia descriptions we
quote are auto-generated by GPT-3.5 and GPT-4o-mini and may be
wrong; the interpretation of those descriptions (e.g.\ deciding
that ``cryptographic keys'' is what feature 17{,}491 detects)
was done by one person in one pass. A careful re-runner should
confirm by inspecting top-activating examples on Neuronpedia
directly rather than trusting the label.

\textbf{Five seeds is a small robustness check.} Control~3 used
$n{=}5$ seeds. The keys-effect range (75--93\%) is consistent
with the headline; a larger sweep would give a tighter interval.
We treat five seeds as sufficient to distinguish ``stable
behavioural effect'' from ``stable headline feature'' but not as
sufficient for a tight effect-size estimate.

\textbf{Position-aggregation choice.} The feature signal we
analyze is the mean SAE encoding across the last three tokens of
the prompt. A finer-grained per-position analysis (which token
position carries the discriminating signal?) is supported by the
pipeline but not reported here.

% ============================================================
\section{Conclusion}
% ============================================================

We took a sparse-feature audit method that has lived mostly in the
mechanistic-interpretability literature and applied it to a fixed
task corpus to ask which features predict task failure. On 300 IOI
prompts with GPT-2 small the method surfaced a striking
single-feature correlate (mean activation $56{\times}$ higher on
failed trials, Neuronpedia label ``cryptographic keys'') and a
clean behavioural fact (failure rate on the ``the keys'' subset is
75--93\% across seeds vs.\ 7.5\% on the other seven objects). Three
controls then revised the reading. The feature fails the ablation;
the raw 768-d residual stream matches the predictive power of the
top hundred features; the identity of the top feature does not
survive a seed change. The behavioural effect is real, but the
mechanistic story we initially attached to it is not. We treat the
gap between what the audit surfaces and what the controls support
as the contribution. The code and data are released. Future
applications of the pattern should report the correlates and the
controls together; the value of the audit is in their joint output,
not in the headline number.

\section*{Reproducibility}

All code, the procedurally-generated 300-prompt corpus, the
greedy-decode prediction sheet, the
$300 \times 24{,}576$ activation matrix, the per-feature
statistics CSV, the ablation per-task sheet, the raw-activation
baseline report, the multi-seed summary, and the figures are
released at \url{https://github.com/mahdinaser/agent-failure-circuits}
under the MIT License. The pipeline runs on Apple Silicon (MPS)
or CPU; no GPU is required for the GPT-2 small experiment. Total
wall-clock on an M3 Max laptop is approximately twenty minutes
for the full $N{=}300$ run plus the per-feature analysis; the
ablation, the raw-baseline run, and the five-seed sweep each take
under ten minutes on the same hardware. Re-running with a
different random seed or extending the corpus to $N > 300$
requires only
\texttt{python src/build\_task\_set.py --n <N> --seed <S>}
followed by the same run-and-analyze commands.

% ============================================================
% Bibliography
% ============================================================

\end{document}